\documentclass[10pt,twocolumn,letterpaper]{article}

\usepackage{cvpr}
\usepackage{times}
\usepackage{epsfig}
\usepackage{graphicx}
\usepackage{amsmath}
\usepackage{amssymb}
\usepackage{setspace}
\usepackage{algorithm}
\usepackage{algorithmic}
\graphicspath{
    {figure/}
}

\usepackage[breaklinks=true,bookmarks=false]{hyperref}

\cvprfinalcopy 

\ifcvprfinal\fi

\begin{document}

\title{HyperNet: Towards Accurate Region Proposal Generation \\ and Joint Object Detection}
\author{Tao Kong$^{1\thanks{This work was done when Tao Kong was an intern at Intel Labs China supervised by Anbang Yao who is responsible for correspondence.}}$ ~~  Anbang Yao$^2$ ~~ Yurong Chen$^2$ ~~ Fuchun Sun$^1$ \\
$^1$State Key Lab. of Intelligent Technology and Systems\\
$^1$Tsinghua National Laboratory for Information Science and Technology (TNList)\\
$^1$Department of Computer Science and Technology, Tsinghua University ~~  
$^2$Intel Labs China\\
\tt\small \{kt14@mails, sunfc@mail\}.tsinghua.edu.cn ~~ \{anbang.yao, yurong.chen\}@intel.com
}
\maketitle

\thispagestyle{empty}

\begin{abstract}
   Almost all of the current top-performing object detection networks employ region proposals to guide the search for object instances. State-of-the-art region proposal methods usually need several thousand proposals to get high recall, thus hurting the detection efficiency. Although the latest Region Proposal Network method gets promising detection accuracy with several hundred proposals, it still struggles in small-size object detection and precise localization (e.g., large IoU thresholds), mainly due to the coarseness of its feature maps. In this paper, we present a deep hierarchical network, namely HyperNet, for handling region proposal generation and object detection jointly. Our HyperNet is primarily based on an elaborately designed Hyper Feature which aggregates hierarchical feature maps first and then compresses them into a uniform space. The Hyper Features well incorporate deep but highly semantic, intermediate but really complementary, and shallow but naturally high-resolution features of the image, thus enabling us to construct HyperNet by sharing them both in generating proposals and detecting objects via an end-to-end joint training strategy. For the deep VGG16 model, our method achieves completely leading recall and state-of-the-art object detection accuracy on PASCAL VOC 2007 and 2012 using only 100 proposals per image. It runs with a speed of 5 fps (including all steps) on a GPU, thus having the potential for real-time processing.
\end{abstract}

\section{Introduction}

\begin{figure}[t]
\begin{center}
   \includegraphics[width=0.8\linewidth]{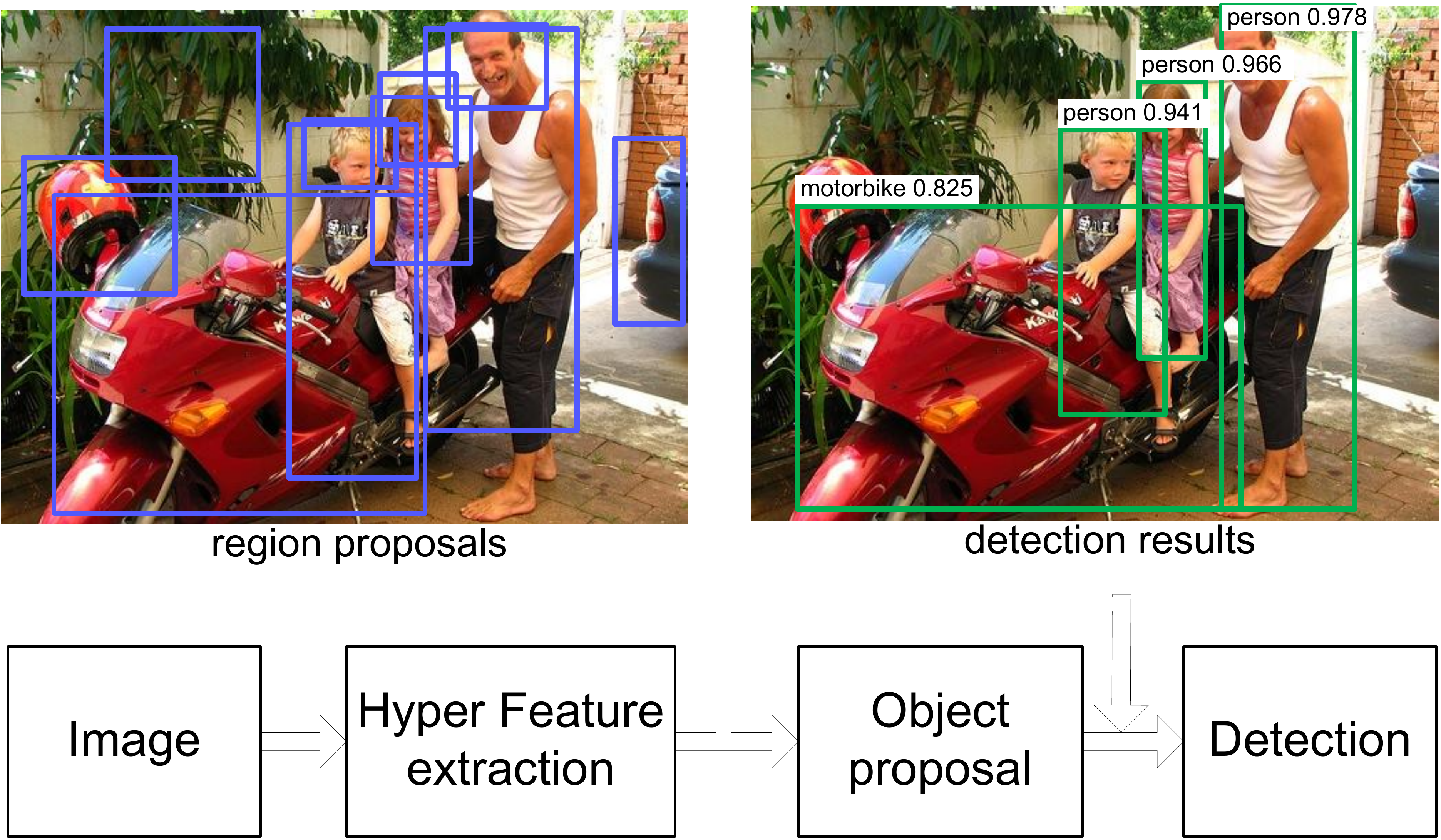}
\end{center}

   \caption{HyperNet object detection overview. \textbf{Topleft}: top 10 object proposals generated by the network.  \textbf{Topright}: detection results with precision value. \textbf{Down}: object proposal generation and detection pipeline. }
\label{pipl}
\vskip -0.2 in
\end{figure}

Generic object detection methods are moving from dense sliding window approaches to sparse region proposal framework. High-quality and category-independent object proposals reduce the number of windows each classifier needs to consider, thus promoting the development of object detection. Most recent state-of-the-art object detection methods adopt such pipeline \cite{fasterrcnn}\cite{girshick2014rich}\cite{sppnet}\cite{multiregion}\cite{bayesian}.  A pioneering work is regions with convolutional neural network (R-CNN)\cite{girshick2014rich}. It first extracts $\sim$2k region proposals by Selective Search \cite{van2011segmentation} method and then classifies them with a pre-trained convolutional neural network (CNN). By employing an even deeper CNN model (VGG16 \cite{simonyan2014very}), it gives 30\% relative improvement over the best previous result on PASCAL VOC 2012 \cite{pascalvoc}.

There are two major keys to the success of the R-CNN: (a) It replaces the hand-engineered features like HOG \cite{dalal2005histograms} or SIFT \cite{sift} with high level object representations obtained from CNN models. CNN features are arguably more discriminative representations. (b) It uses a few thousands of category-independent region proposals to reduce the searching space for an image. One may note that R-CNN relies on region proposals generated by Selective Search. Selective Search takes about 2 seconds to compute proposals for a typical 500$\times$300 image. Meanwhile, feature computation in R-CNN is time-consuming, as it repeatedly applies the deep convolutional networks to thousands of warped region proposals per image \cite{sppnet}.

Fast R-CNN \cite{frcnn} has significantly improved the efficiency and accuracy of R-CNN. Under Fast R-CNN, the convolutional layers are pooled and reused. The region of interest (ROI) pooling strategy allows for extraction of high level feature on proposal windows much faster. Nevertheless, one main issue of Fast R-CNN is that it relies on Selective Search. The region proposal generation step consumes as much running time as the detection network. Another issue of Fast R-CNN is that the last layer output of a very deep CNN is too coarse. So it resizes the image''s short size to 600. In this case, a 32$\times$32 object will be just 2$\times$2 when it goes to the last convolutional layer of VGG16 \cite{simonyan2014very} network. The feature map size is too coarse for classification of some instances with small size. Meanwhile, neighboring regions may overlap each other seriously. This is the reason why Fast R-CNN struggles with small objects on PASCAL VOC datasets.

Recently proposed Region Proposal Network (RPN, also known as Faster R-CNN) combines object proposal and detection into a unified network \cite{fasterrcnn}. The authors add two additional convolutional layers on top of traditional ConvNet output to compute proposals and share features with Fast R-CNN. Using 300 region proposals, RPN with Fast R-CNN produces detection accuracy better than the baseline of Selective Search with Fast R-CNN. However, because of the poor localization performance of the deep layer, this method still struggles with small instances and high IoU thresholds (e.g.,$>$ 0.8)\cite{deepboxes}. Moreover, fewer proposals not only reduce running time but also make detection more accuracy. A proposal generator that can guarantee high recall with small number(e.g., 50) of region boxes is required for better object detection system and other relevant applications \cite{tracking}\cite{saliency}.

Issues in Fast R-CNN and RPN indicate that (a) Features for object proposal and detection should be more informative and (b) The resolution of the layer pre-computed for proposal generation or detection should be reasonable. The deep convolutional layers can find the object of interest with high recall but poor localization performance due to the coarseness of the feature maps. While the low layers of the network can better localize the object of interest but with a reduced recall \cite{deepboxes}. A good object proposal/detection system should combine the best of both worlds.

Recently, Fully Convolution Network (FCN) is demonstrated impressive performance on semantic segmentation task \cite{fcn}\cite{hypercol}. In \cite{fcn}, the authors combine coarse, high layer information with fine, low layer information for semantic segmentation. In-network upsampling enables pixelwise prediction and learning. Inspired by these works, we develop a novel Hyper Feature to combine deep, coarse information with shallow, fine information to make features more abundant.  Our hypothesis is that the information of interest is distributed over all levels of the convolution network and should be well organised. To make resolution of the Hyper Feature appropriate, we design different sampling strategies for multi-level CNN features.

One of our motivations is to reduce the region proposal number from traditional thousands level to one hundred level and even less. We also propose to develop an efficient object detection system. Efficiency is an important issue so that the method can be easily involved in real-time and large-scale applications.

In this paper, we present HyperNet for accurate region proposal generation and joint object detection as shown in Figure \ref{pipl}. We demonstrate that proper fusion of coarse-to-fine CNN features is more suitable for region proposal generation and detection. Our main results are:

\begin{itemize}
\item On object proposal task, our network achieves 95\% recall with just 50 proposals and 97\% recall with 100 proposals, which is significantly better than other existing top methods.
\item On the detection challenges of PASCAL VOC 2007 and 2012, we achieve state-of-the-art mAP of 76.3\% and 71.4\%, outperforming the seminal Fast R-CNN by 6 and 3 points, correspondingly.
\item Our speeding up version can guarantee object proposal and detection accuracy almost in real-time, with 5 fps using very deep CNN models.
\end{itemize}

\section{Related Work}

In this section, we review existing object proposal and detection methods most related to our work, especially deep leaning based methods.

Object proposals \cite{bing}\cite{deepbox}\cite{edgeboxes}\cite{mincut} considerably reduce the computation compared with sliding window methods \cite{dpm}\cite{dpm1} in detection framework. These methods can be classified into two general approaches: traditional methods and deep learning based methods. Traditional methods attempt to generate region proposals by merging multiple segments or by scoring windows that are likely be included in objects. These methods unusually adopt cues like superpixels \cite{van2011segmentation}, edges \cite{edgeboxes}\cite{bing}, saliency \cite{objectness} and  shapes \cite{mcg}\cite{midcut} as features . Recently, some researchers are using CNN to generate region proposals. Deepbox \cite{deepbox} is trained with a slight ConvNet model that learns to re-rank region proposals generated by EdgeBoxes \cite{edgeboxes}. RPN \cite{fasterrcnn} has joined region proposal generator with classifier in one stage or two stages. Both region proposal generation and detection results are promising. In DeepProposal \cite{deepboxes}, a coarse-to-fine cascade on multiple layers of CNN features is designed for generating region proposals.

\begin{figure*}
\begin{center}
    \includegraphics[width=0.9\linewidth]{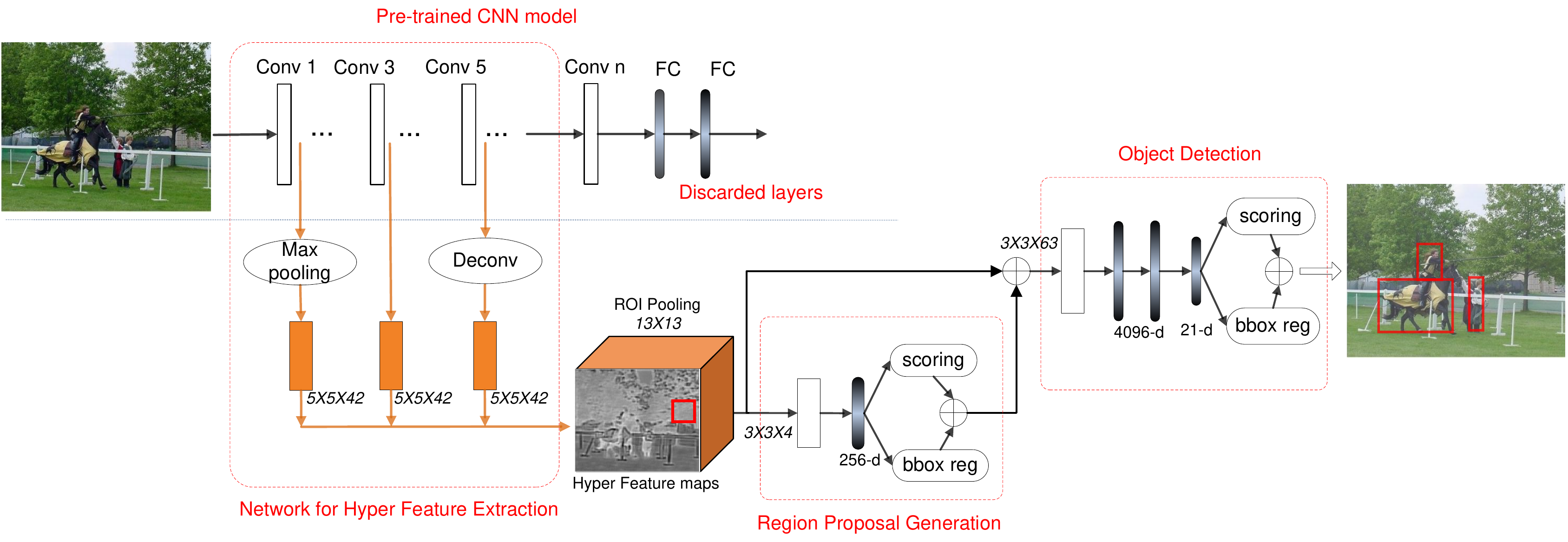}
\end{center}
\vskip -0.1 in
   \caption{HyperNet object detection architecture. Our system (1)
takes an input image, (2) computes Hyper Feature representation, (3) genrates 100 proposals and (4) classifies and makes adjustment for each region.}
\label{framework}
\vskip -0.2 in
\end{figure*}

Object detection aims to localize and recognize every object instance with a bounding box \cite{pascalvoc}\cite{imagenet}. The DPM \cite{dpm} and its variants \cite{dpm2}\cite{dpm3} have been the dominating methods for years. These methods use image descriptors such as HOG \cite{dalal2005histograms}, SIFT \cite{sift}, and LBP \cite{hog} as features and sweep through the entire image to find regions with a class-specific maximum response. With the great success of the deep learning on large scale object recognition \cite{alexnet}, several works based on CNN have been proposed\cite{overfeat}\cite{bayesian}\cite{frcnn}. Girshick et al. \cite{girshick2014rich} propose R-CNN. In this framework, a few thousand category-independent region proposals are adopted for object detection. They also develop a fast version with higher accuracy and speed \cite{sppnet}\cite{frcnn}. Spyros et al. \cite{multiregion} build a powerful localization system based on R-CNN pipeline. They add semantic segmentation results to enhance localization accuracy. In MultiBox \cite{multibox}, region proposals are generated from a CNN model. Different from these works, Redmon et al. \cite{yolo} propose a You Only Look Once (YOLO) framework that predicts bounding boxes and class probabilities directly from full images. Among these methods, R-CNN, Fast-RCNN and MultiBox are proposal based methods. YOLO is proposal free method. In practice, proposal based methods completely outperform proposal free methods with respect to detection accuracy. Some methods share similarities with our work, and we will discuss them in more detail in Section \ref{comp}.

\section{HyperNet Framework}

Our HyperNet framework is illustrated in Figure \ref{framework}. Initially, an entire image is forwarded through the convolutional layers and the activation maps are produced. We aggregate hierarchical feature maps and then compress them into a uniform space, namely Hyper Feature. Next, a slight region proposal generation network is constructed to produce about 100 proposals. Finally, these proposals are classified and adjusted based on the detection module.

\subsection{Hyper Feature Production}

Given an image, we apply the convolutional layers of a pre-trained model to compute feature maps of the entire image. As Fast R-CNN, we keep the image's aspect ratio and resize the short side to 600. Because of subsampling and pooling operations in CNN, these feature maps are not at the same resolution. To combine multi-level maps at the same resolution, we carry out different sampling strategies for different layers.
We add a max pooling layer on the lower layer to carry out subsampling. For higher layers, we add a deconvolutional operation (Deconv) to conduct upsampling. A convolutional layer (Conv) is applied to each sampled result. The Conv operation not only extracts more semantic features but also compresses them into a uniform space.
Finally, we normalize multiple feature maps using local response normalization (LRN)\cite{caffe} and concatenate them to one single output cube, which we call Hyper Feature.

Hyper Feature has several advantages: (a) Multiple levels' abstraction. Inspired by neuroscience, reasoning across multiple levels has been proven beneficial in some computer vision problems \cite{hypercol}\cite{sppnet}. Deep, intermediate and shallow CNN features are really complementary for object detection task as shown in experiments. (b) Appropriate resolution. The feature map resolution for a resized 1000$\times$600 image will be 250$\times$150, which is more suitable for detection. (c) Computation efficiency. All features can be pre-computed before region proposal generation and detection module. There is no redundant computation.

\subsection{Region Proposal Generation}

Designing deep classifier networks on top of feature extractor is as important as the extractor itself. Ren et al. \cite{noc} show that a ConvNet on pre-computed feature maps performs well. Following their findings, we design a lightweight ConvNet for region proposal generation. This ConvNet includes a ROI pooling layer, a  Conv layer and a Fully Connect (FC) layer, followed by two sibling output layers. For each image, this network generates about 30k candidate boxes with different sizes and aspect ratios.

The ROI pooling performs dynamic max pooling over $w\times h$ output bins for each box. In this paper, both $w$ and $h$ are set to 13 based on the validation set. On top of the ROI pooling output, we add two additional layers. One encodes each ROI position into a more abstract feature cube (13$\times$13$\times$4) and the other encodes each cube into a short feature vector (256-d). This network has two sibling output layers for each candidate box. The scoring layer computes the possibility of an object's existence and the bounding box regression layer outputs box offsets.

After each candidate box is scored and adjusted, some region proposals highly overlap each other. To reduce redundancy, we adopt greedy non-maximum suppression (NMS) \cite{girshick2014rich} on the regions based on their scores. For a box region, this operation rejects another one if it has an intersection-over-union (IoU) overlap higher than a given threshold. More concretely, we fix the IoU threshold for NMS at 0.7, which leaves us about 1k  region proposals per image. After NMS, we select the top-k ranked region proposals for detection. We train the detection network using top-200 region proposals, but evaluate different numbers at test time.

\subsection{Object Detection}

The simplest way to implement object detection is to take the FC-Dropout-FC-Dropout pipeline \cite{frcnn}\cite{fasterrcnn}\cite{noc}. Based on this pipeline, we make two modifications.
(a) Before FC layer, we add a Conv layer (3$\times$3$\times$63) to make the classifier more powerful. Moreover, this operation reduces half of the feature dimensions, facilitating following computation.
(b) The dropout ratio is changed from 0.5 to 0.25, which we find is more effective for object classification.
As the proposal generation module, the detection network also has two sibling output layers for each region box. The difference is that there are \emph{N}+1 output scores and 4$\times$\emph{N} bounding box regression offsets for each candidate box (where \emph{N} is the number of object classes, plus 1 for background).

Each candidate box is scored and adjusted using the output layers. We also add a class specific NMS to reduce redundancy. This operation suppresses few boxes, as most boxes have been filtered at the proposal generation step.

\subsection{Joint Training}

For training proposals, we assign a binary class label (of being an object or not) to each box. We assign positive label to a box that has an IoU threshold higher than 0.45 with any ground truth box. We assign negative label to a box if its IoU threshold is lower than 0.3 with all ground truth boxes. We minimize a multi-task loss function.
\begin{equation}
L(k,k^*,t,t^*) = L_{cls}(k,k^*) + \lambda  L_{reg} (t,t^*)
\end{equation}
where the classification loss $L_{cls}$ is Softmax loss of two classes. And the second task loss $L_{reg}$ is bounding box regression for positive boxes. $k^*$ and $k$ are the true and predicted label separately. $L_{reg}(t,t^*) = R(t-t^*)$ where $R$ is the smoothed $L_1$ loss defined in \cite{frcnn}. At proposal generation step, we set regularization $\lambda = 3$ , which means that we bias towards better box locations. At detection step, we optimize scoring and bounding box regression losses with the same weight. $t = (t_x,t_y,t_w,t_h)$ and a predicted vector $t^* = (t_x^*,t_y^*,t_w^*,t_h^*)$  are for positive boxes. We use the parameterizations for $t$ given in R-CNN.
\begin{equation}
\begin{split}
&t_x=(G_x-P_x)/P_w \quad t_y=(G_y-P_y)/P_h\\
&t_w=\log (G_w/P_w) \quad\quad t_h=\log (G_h/P_h)
\end{split}
\end{equation}
where $P^i = (P_x, P_y,P_w,P_h)$ specifies the pixel coordinates of the center of proposal $P$'s bounding box together with $P$'s width and height in pixels. Each ground-truth bounding box $G$ is specified in the same way.

It is not an easy story to design an end-to-end network that includes both region proposal generation and detection, and then to optimize it jointly with back propagation. For detection, region proposals must be computed and adjusted in advance. In practice, we develop a 6-step training process for joint optimization as shown in Algorithm \ref{Trainprocess}.
\begin{algorithm}[h]
\caption{HyperNet training process. After 6 steps, the proposal and detection modules form a unified network.}
\label{Trainprocess}
\begin{algorithmic}
\STATE \textbf{Step 1}: Pre-train a deep CNN model for initializing basic layers in \textbf{Step 2} and \textbf{Step 3}.
\STATE \textbf{Step 2}: Train HyperNet for region proposal generation.
\STATE \textbf{Step 3}: Train HyperNet for object detection using region proposals obtained from \textbf{Step 2}.
\STATE \textbf{Step 4}: Fine-tune HyperNet for region proposal generation sharing Hyper Feature layers trained in \textbf{Step 3}.
\STATE \textbf{Step 5}: Fine-tune HyperNet for object detection using region proposals obtained from \textbf{Step 4}, with shared Hyper Feature layers fixed.
\STATE \textbf{Step 6}: Output the unified HyperNet jointly trained in \textbf{Step 4} and \textbf{Step 5} as the final model.
\end{algorithmic}
\end{algorithm}

Before step 4, object proposal and detection networks are trained separately. After fine-tune of step 4 and step 5, both networks share Network for Hyper Feature Extraction module as seen in Figure \ref{framework}. Finally, we combine two separate networks into a unified network. For proposal/detection, we used a learning rate of 0.005 for the first 100k mini-batches, and 0.0005 for the next 50k mini-batches both in training and fine-tuning. At each mini-batch, 64 RoIs were sampled from a image. We used the momentum term weight 0.9 and the weight decay factor 0.0005. The weights of all new layers were initialized with ``Xavier''. In \cite{fasterrcnn}, Ren et al. develop a 4-step training strategy to share Region Proposal Network with Fast R-CNN. However, we train region proposal generation and detection networks with more powerful features. In addition, the detection module is also redesigned.

\subsection{Speeding up}

\begin{figure}[h]
\vskip -0.1 in
\begin{center}
   \includegraphics[width=0.8\linewidth]{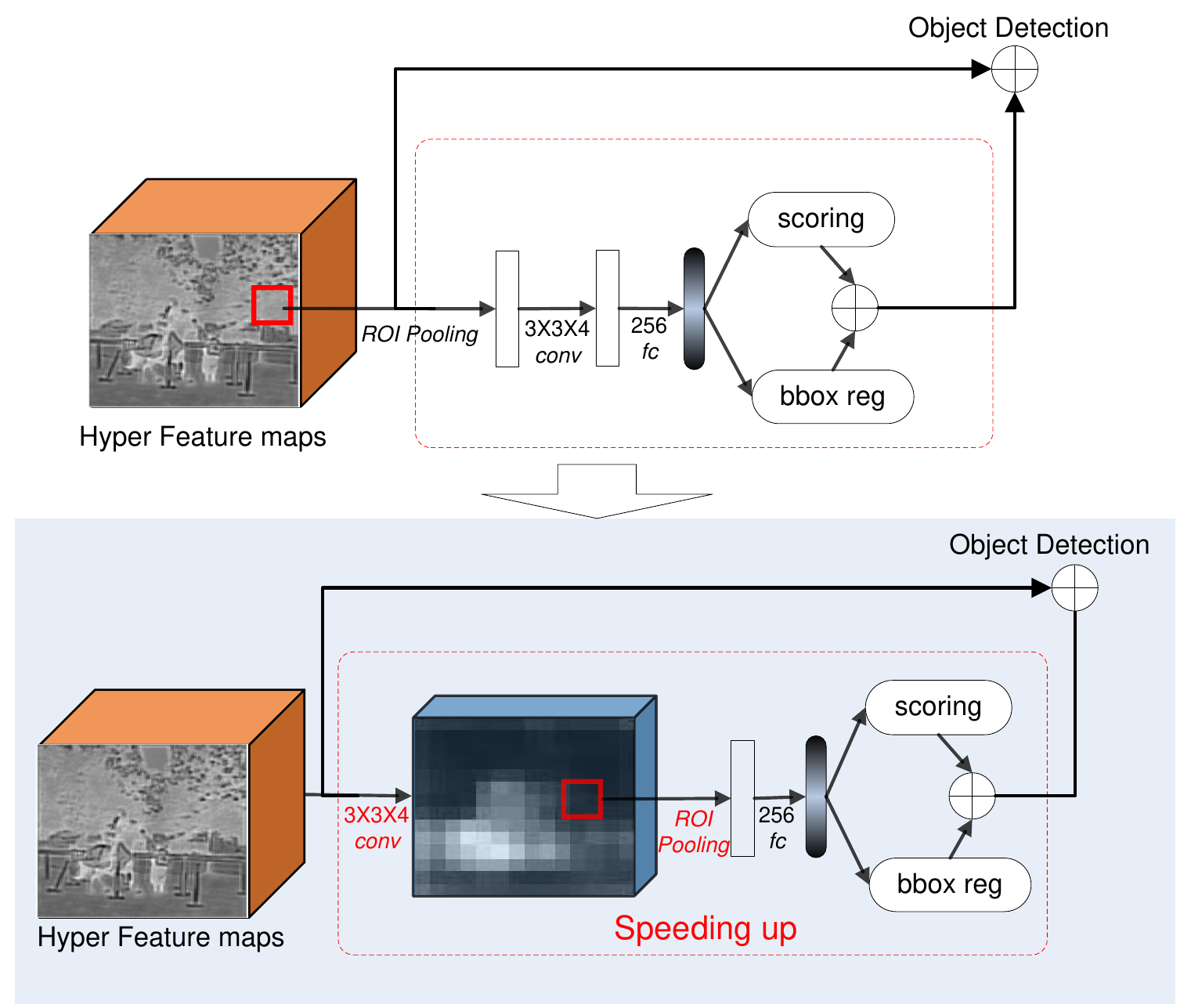}
\end{center}
   \caption{HyperNet speed up. We move the 3$\times$3$\times$4 convolutional layer to the front of ROI pooling to accelerate test speed.}
\label{speedup}
\vskip -0.1 in
\end{figure}

In region proposal generation module, the number of ROIs to be processed is large and most of the forward time is spent in it (about 70\% of the total time). This module needs repeatedly evaluate tens of thousands of candidate boxes as shown in Figure \ref{speedup} top.

Recognizing this fact, we make a minor modification to speed up this process.  As shown in Figure \ref{speedup}, we move the 3$\times$3$\times$4 convolutional layer to the front of ROI pooling layer. This change has two advantages: (a) The channel number of Hyper Feature maps has been significantly reduced (from 126 to 4). (b) The sliding window classifier is more simple (from Conv-FC to FC). Both two characteristics can  speed up region proposal generation process. As we show in experiments, with a little bit drop of recall, the region proposal generation step is almost cost-free (40$\times$ speed up). We also speed up the object detection module with similar changes.

\section{Comparison to Prior Works}\label{comp}

Here, we compare HyperNet with several existing state-of-the-art object proposal and detection frameworks and point out key similarities and differences between them.

\noindent\textbf{Fast R-CNN}\quad Fast R-CNN \cite{frcnn} is the best performing version of R-CNN \cite{girshick2014rich}. HyperNet shares some similarities with Fast R-CNN. Each candidate box predicts a potential bounding box and then scores that bounding box using ConvNet features. However, HyperNet produces object proposal and detection results in a unified network. And the number of region proposals needed is far less than that of Fast R-CNN (100 vs 2000). HyperNet also gets more accurate object detection results.

\noindent\textbf{Faster R-CNN}\quad Unlike Fast R-CNN, the region proposals in Faster R-CNN \cite{fasterrcnn} are produced by RPN. Both Faster R-CNN and the proposed HyperNet have joined region proposal generator with classifier together. Main differences are: (a) Faster R-CNN still relies on Fast R-CNN for object detection while our system unifies region proposal generation and detection into a redesigned network. (b) Our system achieves bounding box regression and region scoring in a different manner. By generating Hyper Feature, our system is more suitable for small object discovery. (c) For high IoU thresholds (e.g.,$>$0.8), our region proposals still perform well.

\noindent\textbf{Deepbox and DeepProposal}\quad Deepbox \cite{deepbox} is a ConvNet model that re-ranks region proposals generated by EdgeBoxes \cite{edgeboxes}. This method follows R-CNN manner to score and refine proposals. Our model, however, firstly computes the feature map of an entire image and then applies detection. DeepProposal \cite{deepboxes} is based on a cascade starting from the last convolutional layer of AlexNet \cite{alexnet}.  It goes down with subsequent refinements until the initial layers of the network. Our network uses in-net sampling to fuse multi-level CNN features. Using 100 region proposals with IoU=0.5, HyperNet gets 97\% recall, 14 points higher than DeepProposal on PASCAL VOC 2007 test dataset.


\noindent\textbf{Multi-region \& Seg-aware}\quad Gidaris et al. \cite{multiregion} propose a multi-region \& semantic segmentation-aware CNN model for object detection. They enrich candidate box representations by additional boxes. They also use semantic segmentation results to enhance localization accuracy. Using these tricks, they get high localization accuracy on PASCAL VOC challenges. However, firstly this method relies on region proposals generated from Selective Search. Secondly, it is time-consuming to evaluate additional boxes and to add semantic segmentation results. In this paper, we propose to develop a unified, efficient, end-to-end training and testing system for proposal generation and detection. HyperNet also gets state-of-the-art object detection accuracy on corresponding benchmarks .

\section{Experimental Evaluation}

We evaluate HyperNet on PASCAL VOC 2007 and 2012 challenges\cite{pascalvoc} and compare results with other state-of-the-art methods, both for object proposal \cite{van2011segmentation}\cite{edgeboxes}\cite{fasterrcnn} and detection \cite{frcnn}\cite{girshick2014rich}. We also provide deep analysis of Hyper Feature affection to object proposal and detection performances.

\subsection{Analysis for Region Proposal Generation}

\begin{figure*}
\centering
\begin{minipage}[t]{0.26\linewidth}
\centering
\includegraphics[width=1\linewidth]{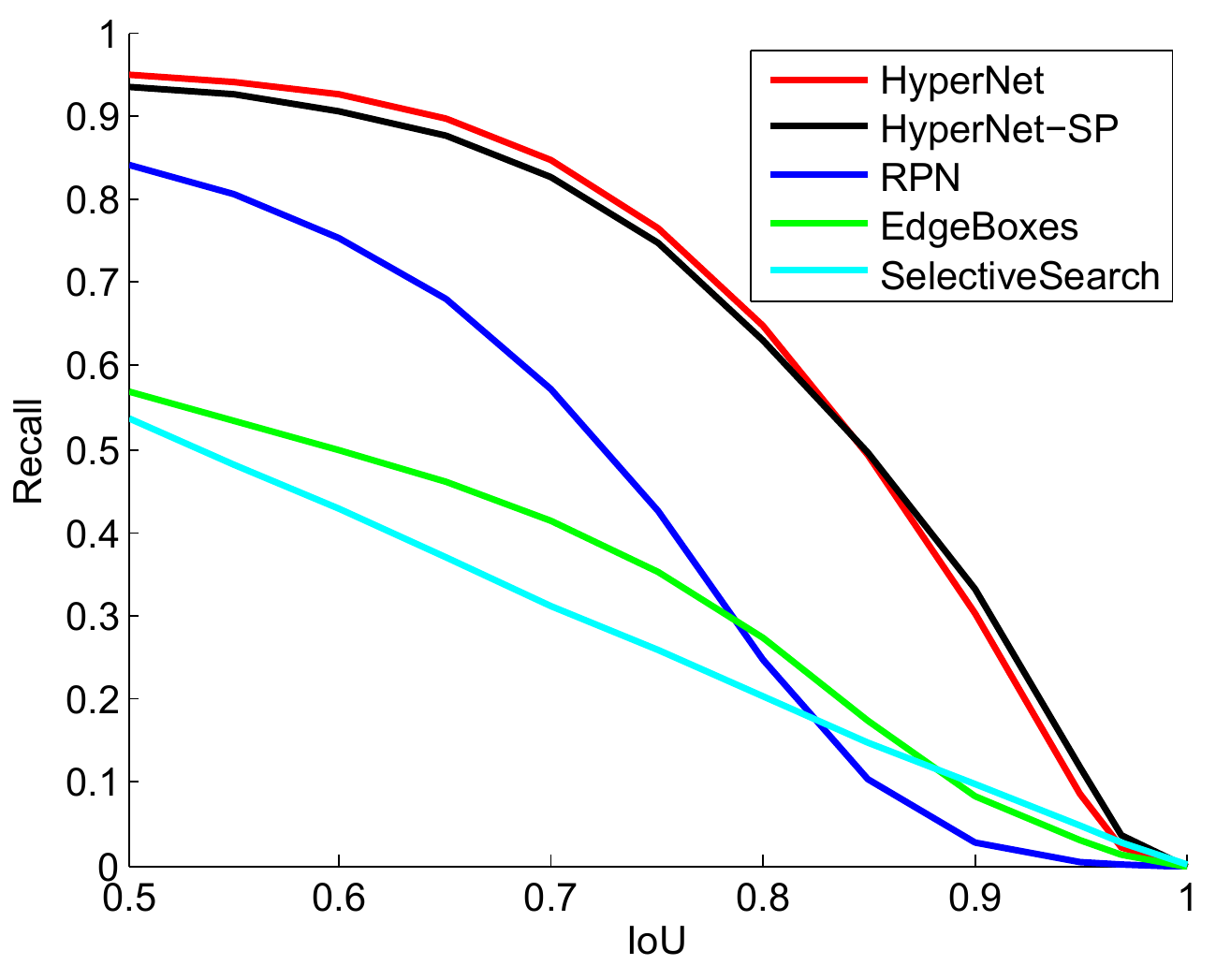}
\end{minipage}
\begin{minipage}[t]{0.26\linewidth}
\centering
\includegraphics[width=1\linewidth]{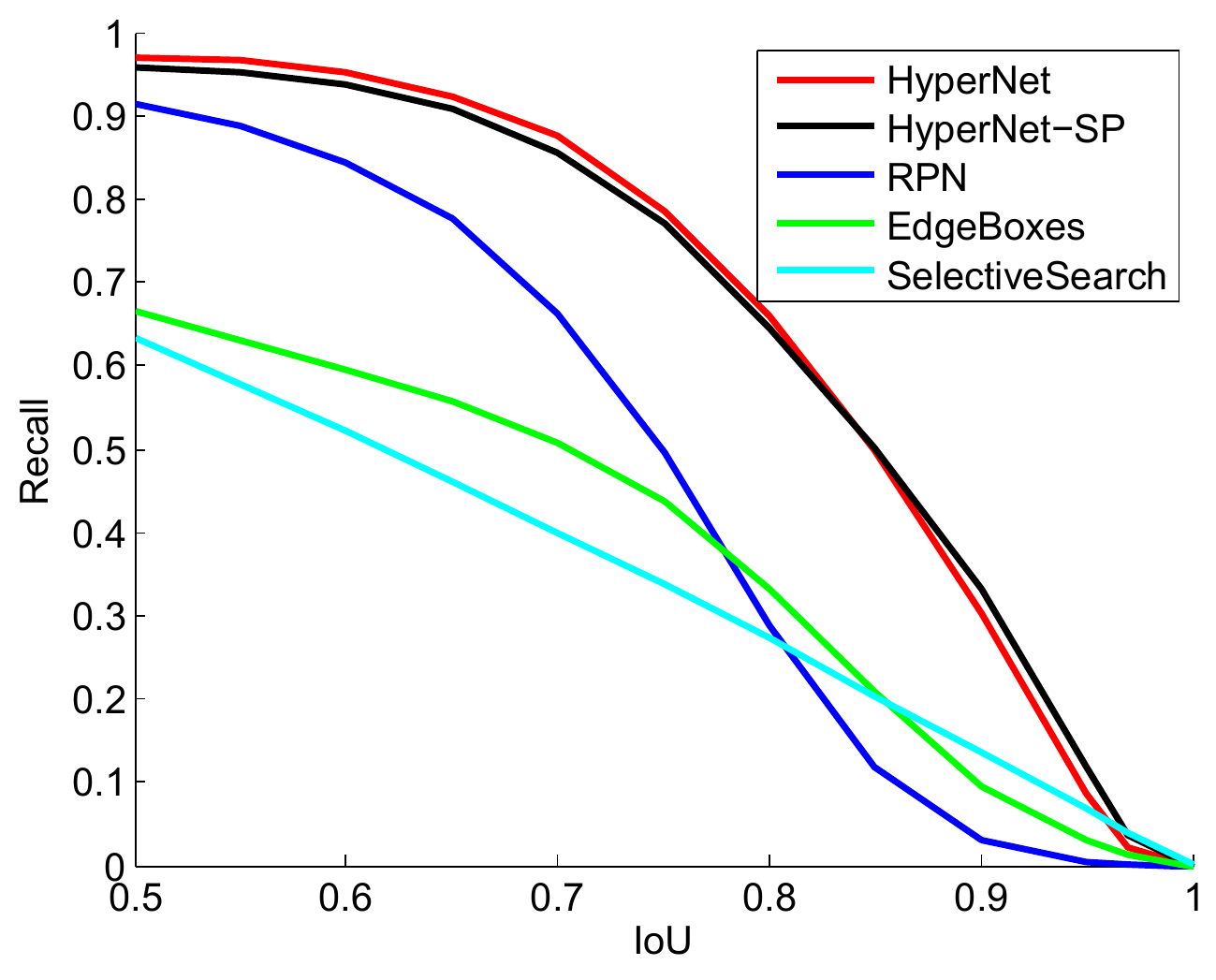}
\end{minipage}
\begin{minipage}[t]{0.26\linewidth}
\centering
\includegraphics[width=1\linewidth]{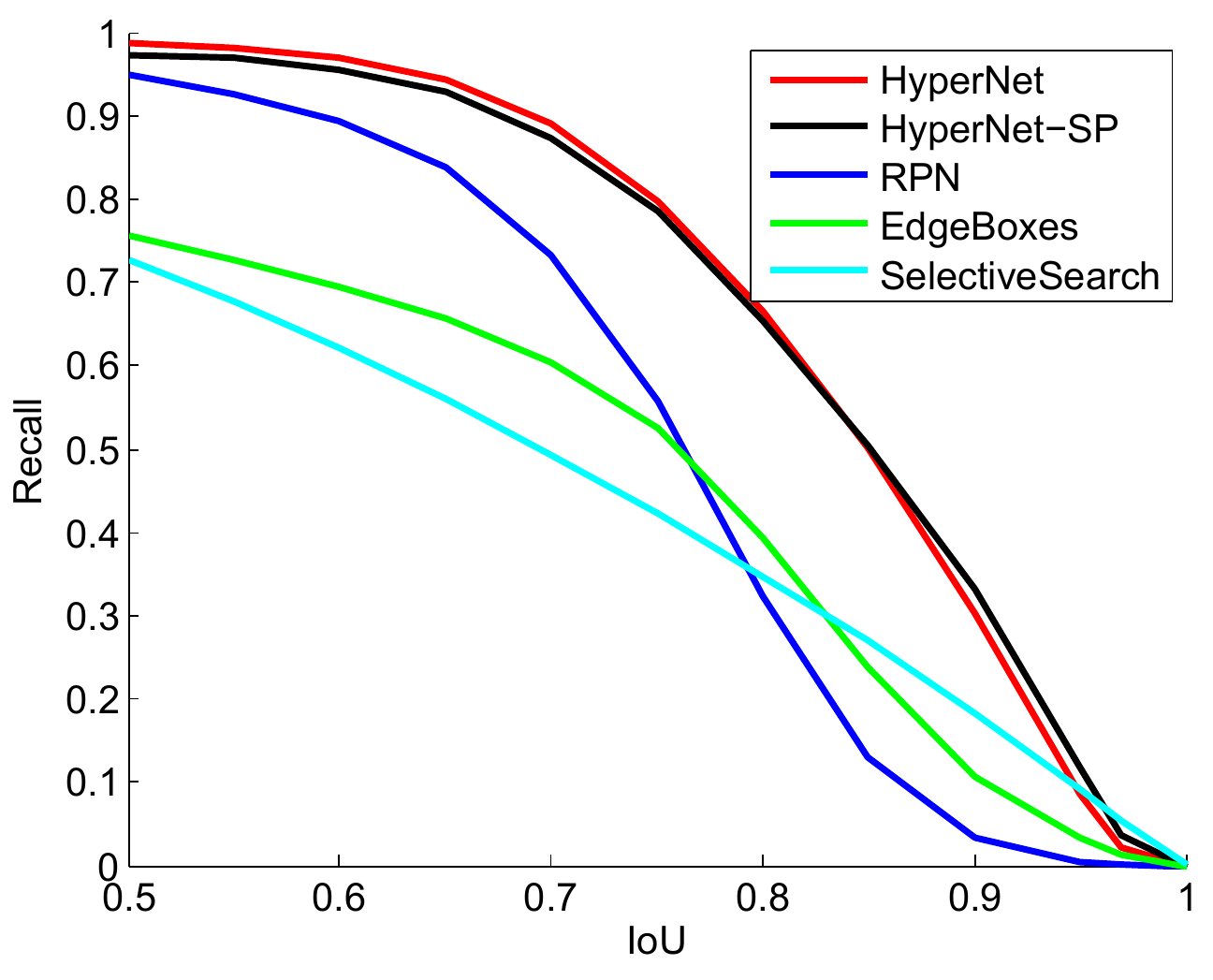}
\end{minipage}
\caption{Recall versus IoU threshold on the PASCAL VOC 2007 test set. \textbf{Left}: 50 region proposals. \textbf{Middle}: 100 region proposals. \textbf{Right}: 200 region proposals.}
\label{iou-recall}
\end{figure*}

\begin{figure*}
\centering
\begin{minipage}[t]{0.26\linewidth}
\centering
\includegraphics[width=1\linewidth]{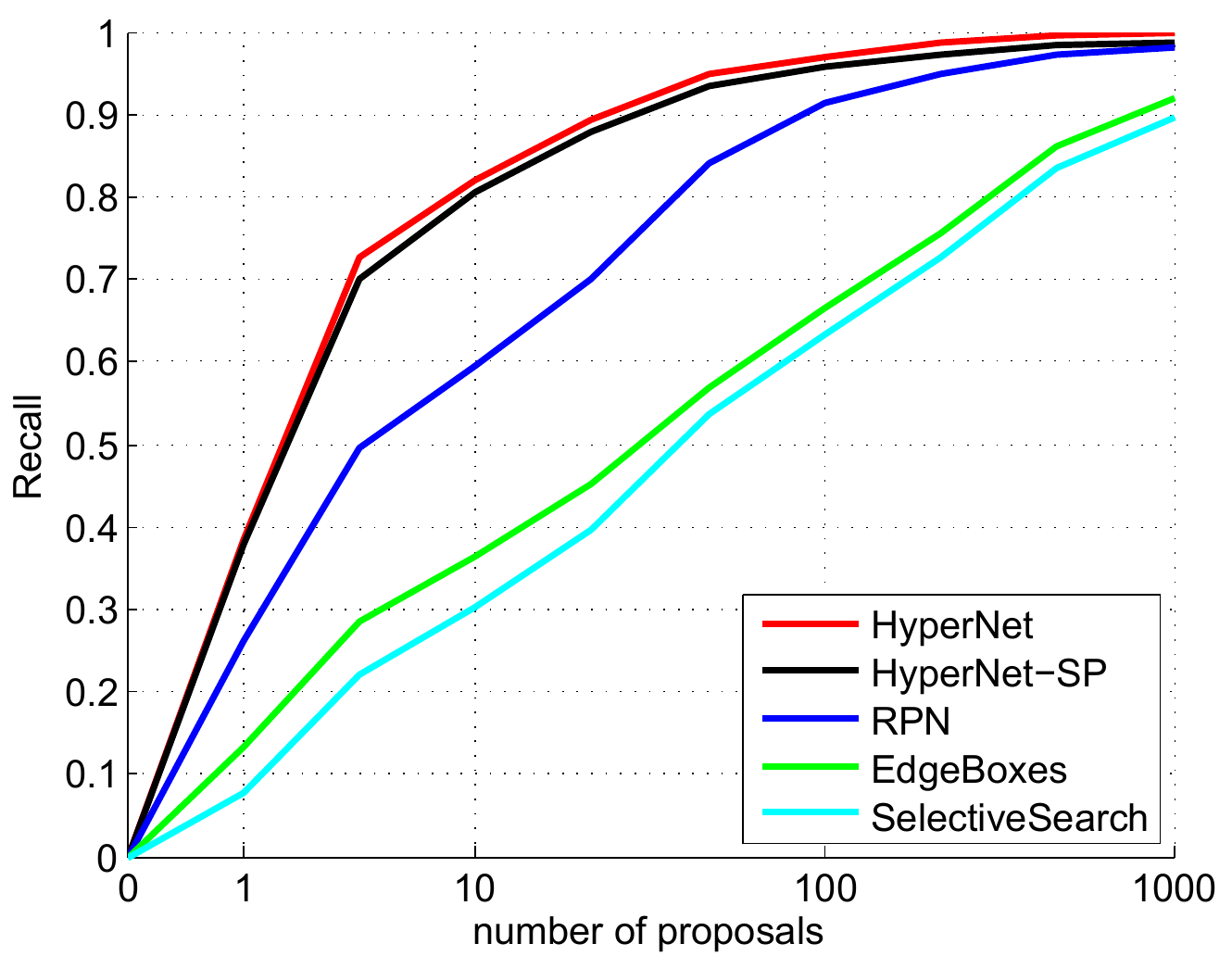}
\end{minipage}
\begin{minipage}[t]{0.26\linewidth}
\centering
\includegraphics[width=1\linewidth]{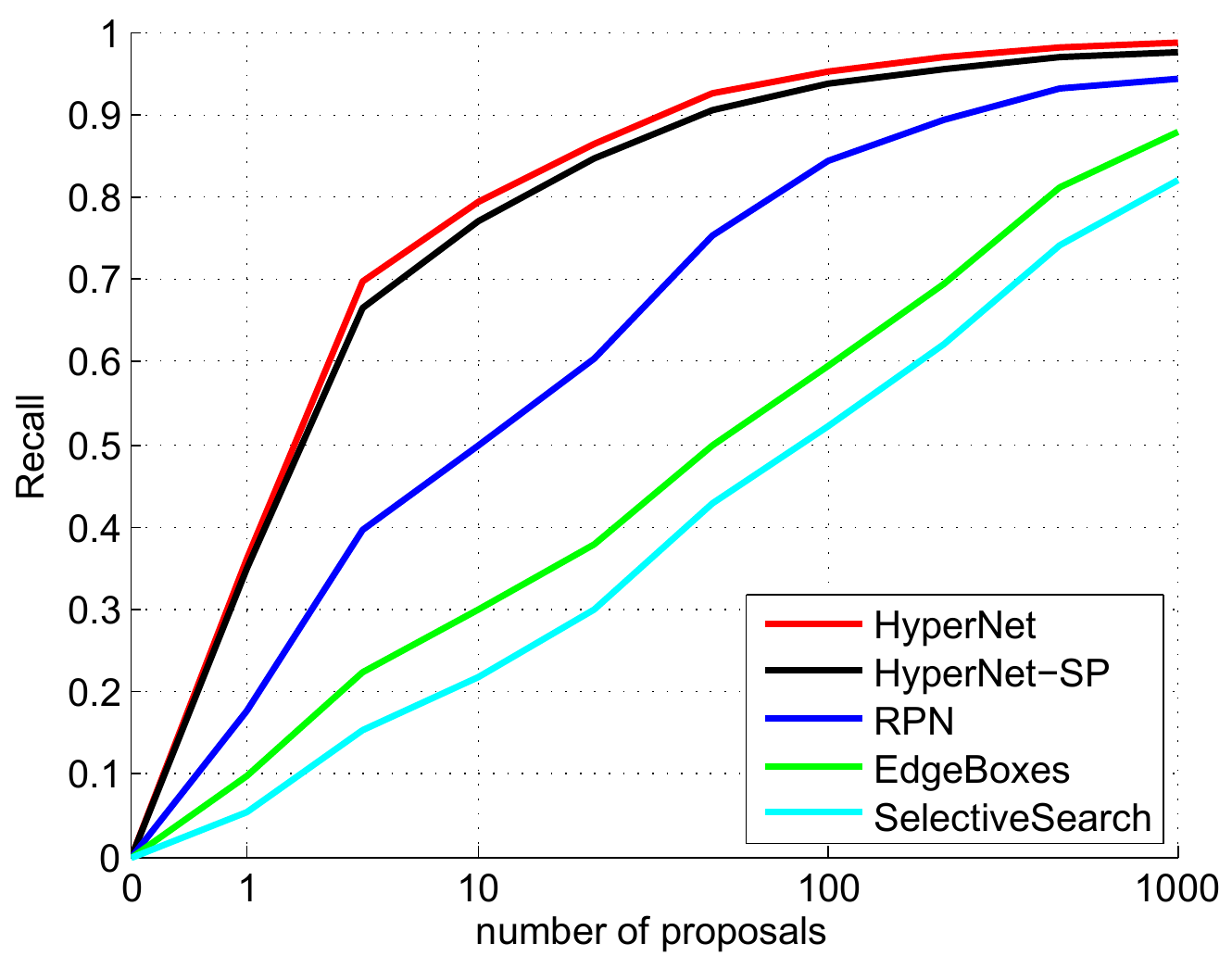}
\end{minipage}
\begin{minipage}[t]{0.26\linewidth}
\centering
\includegraphics[width=1\linewidth]{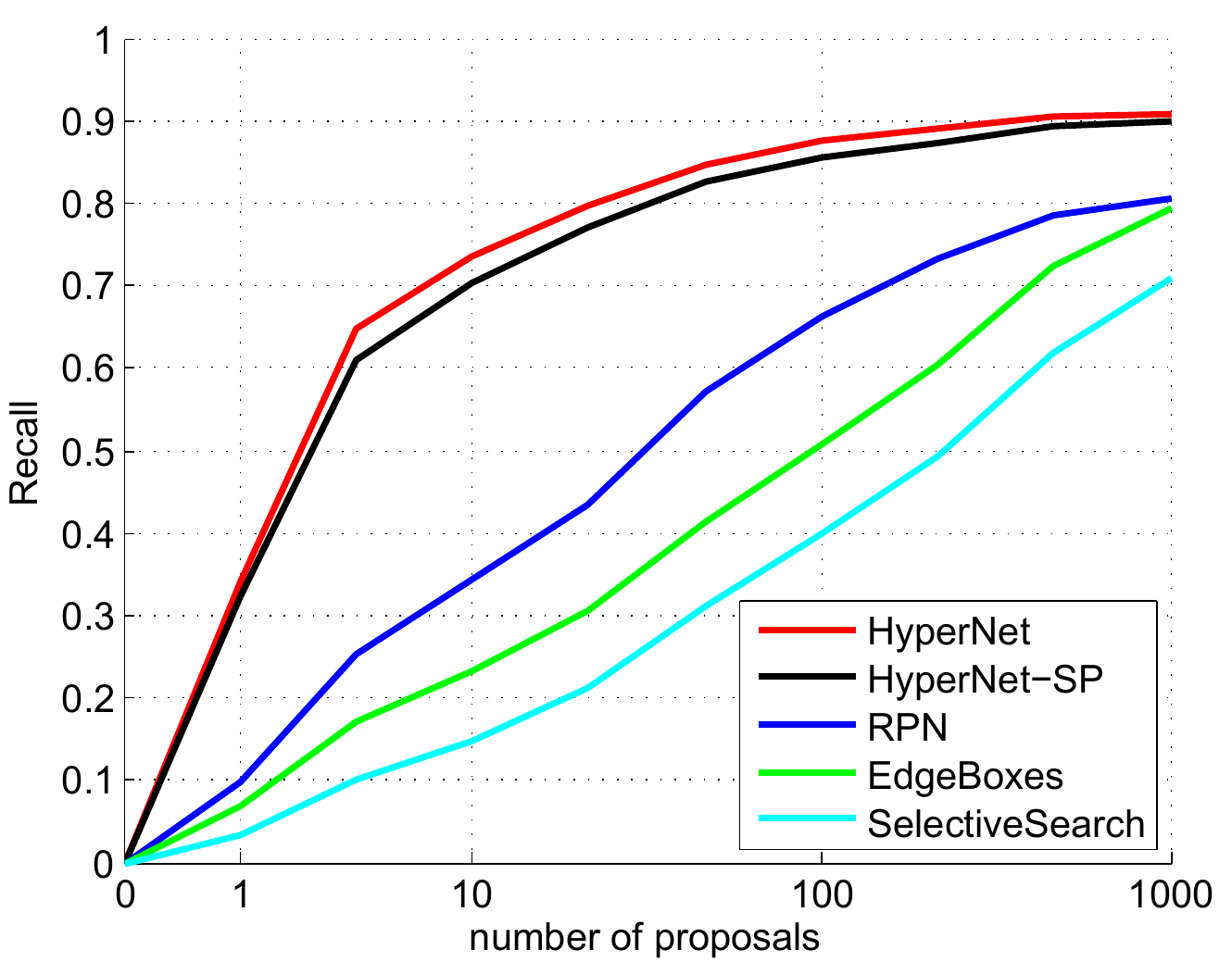}
\end{minipage}
\caption{Recall versus number of proposals on the PASCAL VOC 2007 test set. \textbf{Left}: IoU=0.5. \textbf{Middle}: IoU=0.6. \textbf{Right}: IoU=0.7.}
\label{recall_num_fig}
\vskip -0.1 in
\end{figure*}

In this section, we compare HyperNet against well-known, state-of-the-art object proposal generators. Following \cite{edgeboxes}\cite{van2011segmentation}\cite{fasterrcnn}, we evaluate recall and localization accuracy on PASCAL VOC 2007 test set, which consists of 4,952 images with bounding box annotation for the object instances from 20 categories.

We compare HyperNet with Selective Search, EdgeBoxes and the most recently proposed RPN methods. Curves of recall for methods at different IoU thresholds are plotted. IoU is defined as $\frac{w\cap b}{w\cup b}$ where $b$ and $w$ are the ground truth and object proposal bounding boxes. We evaluate recall vs. overlap for a fixed number of proposals, as shown in Figure \ref{iou-recall}.  The $N$ proposals are the top-$N$ ranked ones based on the confidence generated by these methods.

The plots show that our region proposal generation method performs well when the region number drops from 2k to one hundred level and even less. Specifically, with 50 region proposals, HyperNet gets 95\% recall, outperforming RPN  by 11 points, Selective Search by 42 points and Edgeboxes by 39 points with IoU = 0.5 (Figure \ref{iou-recall} left). Using 100 and 200 region proposals, our network exceeds RPN by 6 and 4 points correspondingly. HyperNet also surpasses Selective Search and EdgeBoxes by a significant margin.

Both RPN and HyperNet achieve promising detection results compared with methods without CNN. However, for high IoU thresholds(e.g., $>$ 0.8), the recall of RPN drops sharply compared with our method.  RPN's features used for regression at an anchor are of the same spatial size (3$\times$3), which means different boxes of scales at the same  position share features. It makes sense with loose IoU (e.g., 0.5). But cannot achieve high recall with strict thresholds \cite{fasterrcnn}. HyperNet achieves good results across a variety of IoU thresholds, which is desirable in practice and plays an important role in object detectors' performance \cite{objectness}.

Figure \ref{recall_num_fig} shows recall versus number of proposals for different methods. Hosang et al. \cite{whatobjectness} show that this criteria correlates well with detection performance. An object proposal with 0.5 IoU threshold is too loose to fit the ground truth object, which usually leads to the failure of later object detectors. In order to achieve good detection results, an object proposal with higher IoU thresholds such as 0.7 is desired. We also show higher IoU threshold results (Table \ref{proposal_num}). Achieving a recall of 75\% requires 20 proposals using HyperNet, 250 proposals using RPN, 800 proposals using EdgeBoxes and 1400 proposals using Selective Search with IoU=0.7.

\begin{table}\small
\begin{center}
\begin{tabular}{l|p{1.3cm}p{1.1cm}p{0.4cm}p{0.8cm}c}

\scriptsize Recall & \scriptsize SelectiveSearch &\scriptsize Edgeboxes &\scriptsize RPN &\scriptsize HyperNet& \scriptsize HyperNet-SP\\
\hline
50\% & \centering 300&\centering	100&\centering	30&\centering	\textbf{5}&7\\
75\% & \centering 1400&\centering	800&\centering	250&\centering	\textbf{20}& 30 \\
\end{tabular}
\end{center}
\vskip -0.1 in
\caption{Region proposal number needed for different recall rate with IoU=0.7}
\label{proposal_num}
\vskip -0.2 in
\end{table}

\subsection{PASCAL VOC 2007 Results}

\begin{table*}\scriptsize \centering
\begin{center}
\begin{spacing}{1.25}
\begin{tabular}{p{2.2cm}|p{0.43cm}|p{0.25cm}p{0.25cm}p{0.25cm}p{0.25cm}p{0.25cm}p{0.25cm}p{0.25cm}p{0.25cm}p{0.25cm}p{0.25cm}p{0.25cm}p{0.25cm}p{0.25cm}p{0.3cm}p{0.3cm}p{0.3cm}p{0.25cm}p{0.25cm}p{0.25cm}c}
\scriptsize Approach&\scriptsize\centering mAP&\centering\scriptsize aero&\centering\scriptsize bike&\centering\scriptsize bird&\centering\scriptsize boat&\centering\scriptsize bottle&\centering\scriptsize bus&\centering\scriptsize car&\centering\scriptsize cat& \centering\scriptsize chair&\centering\scriptsize cow&\centering\scriptsize table&\centering\scriptsize dog&\centering\scriptsize horse&\centering\scriptsize mbike&\centering\scriptsize person&\centering\scriptsize plant&\centering\scriptsize sheep&\centering\scriptsize sofa&\centering\scriptsize train&\scriptsize tv \\
\hline
  Fast R-CNN &70.0& 77.0 & 78.1& 69.3& 59.4& 38.3& 81.6 & 78.6& 86.7 & 42.8& 78.8& 68.9 & 84.7& 82.0& 76.6& 69.9& 31.8& 70.1& 74.8& 80.4 & 70.4 \\
  Faster R-CNN&73.2& 76.5& 79.0& 70.9& 65.5& 52.1& \textbf{83.1}& 84.7& 86.4& 52.0& \textbf{81.9}& 65.7& 84.8& 84.6& 77.5& 76.7& 38.8& 73.6& 73.9& \textbf{83.0}& 72.6 \\
  \hline
  HyperNet(AlexNet)&65.9&70.8&75.2&58.2&57.7&40.5&77.6&76.9&74.9&41.3&71.8&66.9&73.7&79.8&75.9&70.9&35.2&62.4&69.2&74.9&63.6\\
  HyperNet & \textbf{76.3} & \textbf{77.4} &\textbf{83.3}& 75.0& \textbf{69.1}& 62.4& \textbf{83.1}& \textbf{87.4}& \textbf{87.4}& \textbf{57.1}& 79.8& 71.4& \textbf{85.1}& \textbf{85.1}& \textbf{80.0}& \textbf{79.1}& \textbf{51.2}& \textbf{79.1} &\textbf{75.7}& 80.9& \textbf{76.5}\\
  HyperNet-SP & 74.8& 77.3& 82.0& \textbf{75.4}& 64.1& \textbf{63.5}& 82.5& \textbf{87.4}& 86.6& 55.1& 79.3& \textbf{71.5}& 81.4& 84.2& 77.6& 78.4& 45.5& 77.4& 73.2& 78.7& 74.8  \\
\end{tabular}
\end{spacing}
\end{center}
\vskip -0.2 in
\caption{Results on PASCAL VOC 2007 test set (with IoU = 0.5). Rows 3-5 present our HyperNet performance. HyperNet-SP denotes the speeding up version. The entries with the best APs for each object category are bold-faced}
\label{voc07}
\end{table*}

\begin{table*}\scriptsize \centering
\begin{center}
\begin{spacing}{1.25}
\begin{tabular}{p{2.2cm}|p{0.43cm}|p{0.25cm}p{0.25cm}p{0.25cm}p{0.25cm}p{0.25cm}p{0.25cm}p{0.25cm}p{0.25cm}p{0.25cm}p{0.25cm}p{0.25cm}p{0.25cm}p{0.25cm}p{0.3cm}p{0.3cm}p{0.3cm}p{0.25cm}p{0.25cm}p{0.25cm}c}
\scriptsize Approach&\scriptsize\centering mAP&\centering\scriptsize aero&\centering\scriptsize bike&\centering\scriptsize bird&\centering\scriptsize boat&\centering\scriptsize bottle&\centering\scriptsize bus&\centering\scriptsize car&\centering\scriptsize cat& \centering\scriptsize chair&\centering\scriptsize cow&\centering\scriptsize table&\centering\scriptsize dog&\centering\scriptsize horse&\centering\scriptsize mbike&\centering\scriptsize person&\centering\scriptsize plant&\centering\scriptsize sheep&\centering\scriptsize sofa&\centering\scriptsize train&\scriptsize tv \\
\hline
  MR-CNN\cite{multiregion} &36.6& 49.5 & 50.5& 29.2& 23.5& 17.9& 51.3 & 50.4& 48.1 & 20.6& 38.1& 37.5 & 38.7& 29.6& 40.3& 23.9& 15.1& 34.1& 38.9& 42.2 & 52.1 \\
  MR-CNN-Best\cite{multiregion}&48.4& 54.9& 61.3& 43.0& 31.5& 38.3& 64.6& 65.0& 51.2& 25.3& 54.4& 50.5& 52.1& 59.1& 54.0& 39.3& 15.9& 48.5& 46.8& 55.3& 57.3 \\
  \hline
  HyperNet & \textbf{58.2}& 	\textbf{64.9}& 	\textbf{64.7}& 	\textbf{52.8}	& 47.9& 	\textbf{50.6}& 	73.1& 	69.8& 	66.8& 	\textbf{34.1}	& \textbf{61.8}& 	\textbf{53.8}& 	\textbf{61.4}& 	\textbf{66.4}& 	\textbf{56.6}& 	\textbf{57.2}& 	\textbf{28.5}	& 64.8	& 60.0	& \textbf{64.5}	& \textbf{64.4}\\
  HyperNet-SP & 57.9	&62.7	&63.4	&52.9&	\textbf{48.3}&	50.4&	\textbf{75.7}	&\textbf{72.5}	&\textbf{67.4}	&33.5	&59.3&	53.8&	60.0&	64.9	&56.2&	57.2	&26.1&	\textbf{64.9}	&\textbf{60.3}&	64.1	&65.2\\
\end{tabular}
\end{spacing}
\end{center}
\vskip -0.2 in
\caption{Results on PASCAL VOC 2007 test set (with IoU = 0.7). Rows 1-2 present Multi-region \& Seg-aware methods\cite{multiregion} for comparison. Rows 3-4 present our HyperNet performance.}
\label{voc0770}
\vskip -0.2 in
\end{table*}

\begin{table*}[t]\scriptsize \centering
\begin{center}
\begin{spacing}{1.25}
\begin{tabular}{p{2.2cm}|p{0.43cm}|p{0.25cm}p{0.25cm}p{0.25cm}p{0.25cm}p{0.25cm}p{0.25cm}p{0.25cm}p{0.25cm}p{0.25cm}p{0.25cm}p{0.25cm}p{0.25cm}p{0.25cm}p{0.3cm}p{0.3cm}p{0.3cm}p{0.25cm}p{0.25cm}p{0.25cm}c}

 \scriptsize Approach&\scriptsize\centering mAP&\centering\scriptsize aero&\centering\scriptsize bike&\centering\scriptsize bird&\centering\scriptsize boat&\centering\scriptsize bottle&\centering\scriptsize bus&\centering\scriptsize car&\centering\scriptsize cat& \centering\scriptsize chair&\centering\scriptsize cow&\centering\scriptsize table&\centering\scriptsize dog&\centering\scriptsize horse&\centering\scriptsize mbike&\centering\scriptsize person&\centering\scriptsize plant&\centering\scriptsize sheep&\centering\scriptsize sofa&\centering\scriptsize train&\scriptsize tv \\
\hline
  Fast R-CNN &68.4&	82.3&	78.4&	70.8&	52.3&	38.7&	77.8&	71.6&	\textbf{89.3}&	44.2&	73.0&	55.0&	\textbf{87.5}&	80.5&	80.8&	72.0&	35.1&	68.3&	\textbf{65.7}&	80.4&	64.2 \\
  Faster R-CNN &70.4&	\textbf{84.9}&	\textbf{79.8}&	\textbf{74.3}&	53.9&	49.8&	77.5&	75.9&	88.5&	45.6&	\textbf{77.1}&	\textbf{55.3}&	86.9&	\textbf{81.7}&	80.9&	79.6&	40.1&	72.6&	60.9&	\textbf{81.2}&61.5	 \\
  NoC &68.8	&82.8	&79.0	&71.6&	52.3&	\textbf{53.7}	&74.1	&69.0&	84.9&	46.9&	74.3&	53.1&	85.0&	81.3&	79.5&	72.2& 38.9	& 72.4	&59.5	&76.7	&\textbf{68.1}		 \\
\hline
  HyperNet & \textbf{71.4}&	84.2&	78.5&	73.6&	\textbf{55.6}	&\textbf{53.7}&	\textbf{78.7}&	\textbf{79.8}&	87.7&	\textbf{49.6}&	74.9&	52.1&	86.0&	\textbf{81.7}&	\textbf{83.3}&	\textbf{81.8}&	\textbf{48.6}&	\textbf{73.5}&	59.4&	79.9&	65.7\\
  HyperNet-SP & 71.3	&84.1&	78.3	&73.3&	55.5&	53.6&	78.6&	79.6	&87.5&	49.5&	74.9	&52.1&	85.6&	81.6&	83.2&	81.6&	48.4&	73.2&	59.3&	79.7&	65.6 \\

\end{tabular}
\end{spacing}
\end{center}
\vskip -0.2 in
\caption{Results on PASCAL VOC 2012 test set reported by the evaluation server. Rows 4-5 present our HyperNet performance. HyperNet-SP denotes the speeding up version.}
\label{voc12}
\vskip -0.1 in
\end{table*}
We compare HyperNet to Fast R-CNN and Faster R-CNN for generic object detection on PASCAL VOC 2007. This dataset covers 20 object categories, and the performance is measured by mean average precision (mAP) on the test set. All methods start from the same pre-trained VGG16 \cite{simonyan2014very} network and use bounding box regression. We refer to VGG16 based HyperNet if not explain specially.

Fast R-CNN with Selective Search achieves a mAP of 70.0\%. Faster R-CNN's result is 73.2\%. HyperNet achieves a mAP of 76.3\%, 6.3 points higher than Fast R-CNN and 3.1 points higher than Faster R-CNN. As we have shown above, this is because proposals generated by HyperNet are more accurate than Selective Search and RPN. HyperNet is elaborately designed and benefits from more informative Hyper Feature.

Reasonable resolution of Hyper Feature makes for better object localization, especially when the object size is small. For object of small size, our detection network outperforms Faster R-CNN by a significant margin as seen in Table \ref{voc07}. For bottle, HyperNet achieves 62.4\% AP, 10.3 points improvement and for potted plant, HyperNet achieves 51.2\% AP, 12.4 points higher than Faster R-CNN. The speed up version also keeps up effectiveness. Table \ref{voc0770} shows the detection results with IoU = 0.7, we outperform the best result of \cite{multiregion} by about 10 points with respect to mAP.

We also present a small network trained based on the AlexNet architecture\cite{alexnet}, as shown in Table \ref{voc07} (row 3). This network gets a 65.9\% mAP. For small instances such as bottle and potted plant, the detection performance is in comparable with that of the very deep Fast R-CNN model. These results demonstrate that a light weight HyperNet can give excellent performance for small object detection.

\subsection{PASCAL VOC 2012 Results}

We compare against top methods on the comp4 (outside data) track from the public leaderboard on PASCAL VOC 2012. As the data statistics are
similar to VOC 2007, the training data is the union set of all VOC 2007, VOC 2012 train and validation dataset, following \cite{frcnn}. Networks on Convolutional feature maps(NoC) \cite{noc} is based on SPPNet \cite{sppnet}. HyperNet achieves the top result on VOC 2012 with a mAP of 71.4\%  (Table \ref{voc12}). This is 3.0 points and 1.0 points higher than the counterparts. For small objects (`bottle', `chair', and `plant'), our network still outperforms others.  The speed up version also gets state-of-the-art mAP of 71.3\% with efficiency.

\subsection{The Role of Hyper Feature}

An important property of HyperNet is that it combines coarse-to-fine information across deep CNN models. However, does this strategy really help? We design a set of experiments to elucidate this question. Starting from AlexNet, we separately train different models and see their performances. Firstly, we train a single layer for object proposals (layer 1, 3 and 5). Secondly, we combine layer 3 and 5 together and finally, layer 1, 3 and 5 are all assembled to get results. For fairness, feature maps are normalized to the same resolution and all networks are trained with the same configuration.

\begin{figure}[t]
\begin{center}
   \includegraphics[width=0.8\linewidth]{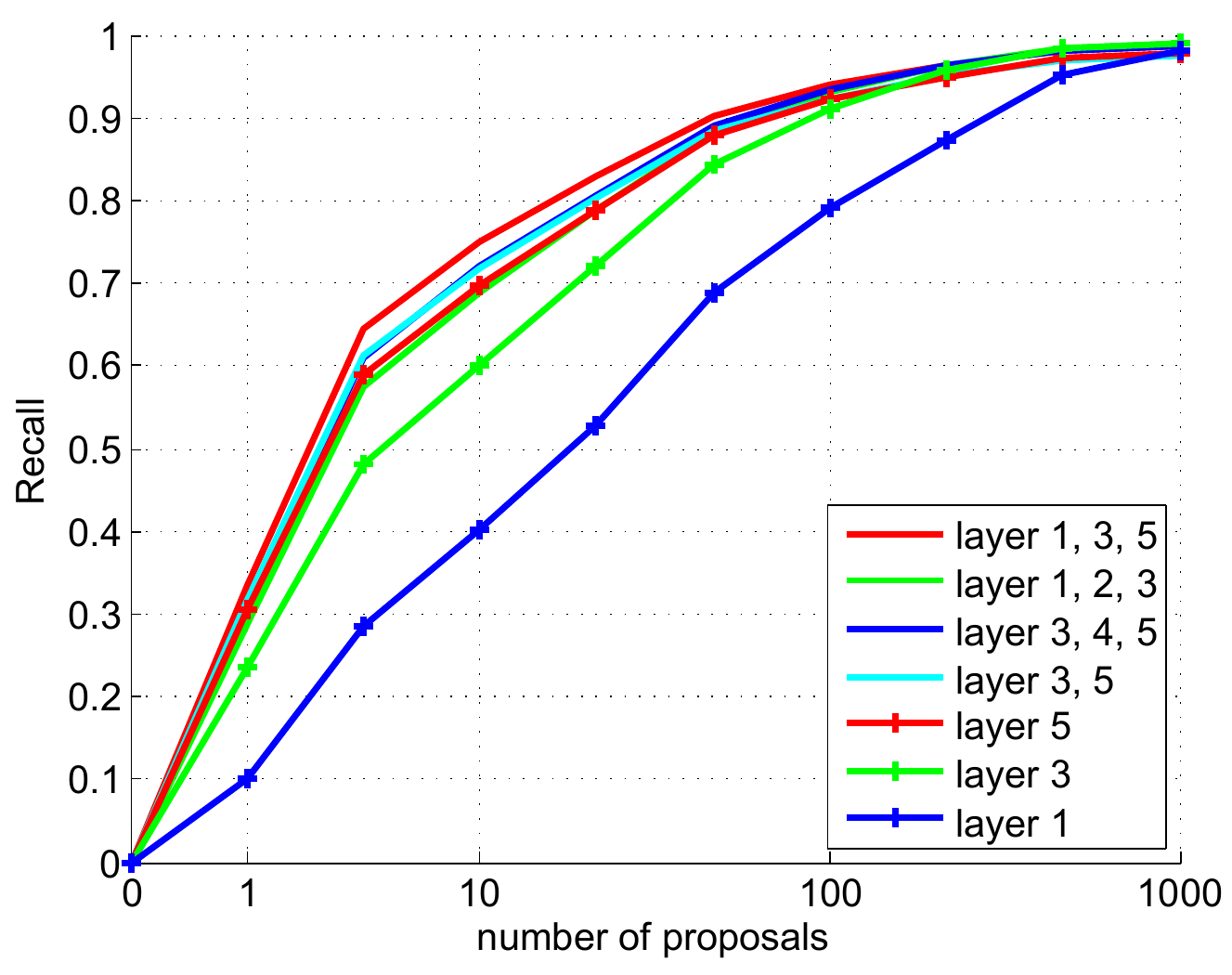}
\end{center}
   \caption{Recall versus number of proposals for different layer combinations using AlexNet (IoU = 0.5).}
\label{caffe_dif_layer}
\vskip -0.2 in
\end{figure}

Unsurprisingly, we find that the combination of layer 1, 3 and 5 works the best, as shown in Figure \ref{caffe_dif_layer}. This result indicates two keys: (a) The multi-layer combination works better than single layer, both for proposal and detection. (b) The last layer performs better than low layers. This is the reason why most systems use the last CNN layer for region proposal generation or detection \cite{deepboxes}\cite{girshick2014rich}\cite{frcnn}. The detection accuracy with respect to mAP is shown in Table \ref{map_comp}.

\begin{table}[t]\small
\begin{center}
\begin{tabular}{c|cc}
Layers&Proposal recall &Detection mAP\\
\hline
1&82.15\%&62.8\%\\
3&93.19\%&63.8\%\\
5&94.98\%&64.2\%\\
\hline
3+5&95.00\%&64.4\%\\
1+2+3&94.79\% &63.8\%\\
3+4+5&95.43\%&64.7\%\\
1+3+5&\textbf{96.16\%}&\textbf{65.9\%}\\
\end{tabular}
\end{center}
\caption{Proposal and detection performance with different layer combination strategies. The region proposal number is 100 for evaluation (IoU = 0.5).}
\label{map_comp}
\vskip -0.2 in
\end{table}

\subsection{Combine Which Layers?}

Hyper Feature is effective for region proposal generation and detection, mainly because of its richness and appropriate resolution. But it also raises another question: which layers should we combine to get the best performance?

To answer this question, we train three models based on AlexNet. The first model combines layer 1, 3 and 5. The second network combines layer 1, 2 and 3 and the final model combines layer 3, 4 and 5. In this section, all networks are trained with the same configuration.

Figure \ref{caffe_dif_layer} shows region proposal performances for different models. There is no sharp difference within these results. However, combining layer 1, 3 and 5 outperforms other networks. Because adjacent layers are strongly correlated, combinations of low layers or high layers behave not that excellent. This indicates that the combination of wider coarse-to-fine CNN features is more important.

We evaluate the detection performance on PASCAL VOC 2007 for these models (see Table \ref{map_comp}). Combining layer 1, 3 and 5 also gets the best detection result (mAP=65.9\%). These detection results demonstrate the effectiveness of the low-to-high combination strategy.

\subsection{Hyper Feature Visualization}

\begin{figure}[t]
\begin{center}
   \includegraphics[width=0.9\linewidth]{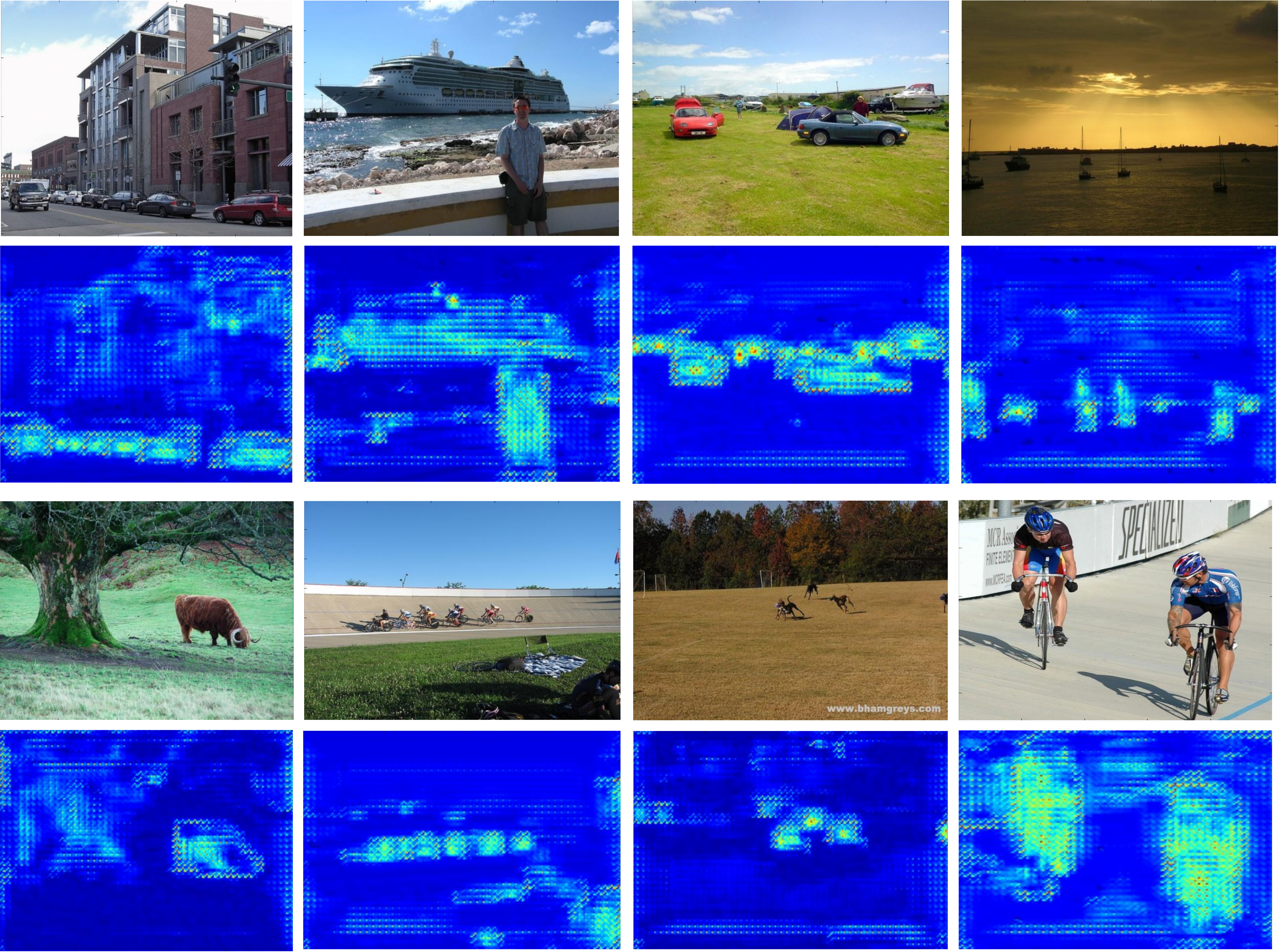}
\end{center}
   \caption{Hyper Feature visualization. Row 1 and 3: input images. Row 2 and 4: corresponding Hyper Feature maps}
\label{vis_fig}
\end{figure}

Figure \ref{vis_fig} shows visualizations for Hyper Features. The feature maps involve not only the strength of the responses, but also their spatial positions. We can see that the feature maps have the potentiality of projecting objects. The area with obvious variation in visualization is more likely to be or part of an object with interest. For example, the particular feature map focuses on cars, but not the background buildings in the first picture. These objects in the input images activate the feature maps at the corresponding positions.

\subsection{Running Time}

We evaluate running time for methods on PASCAL VOC 2007 test dataset, as shown in Table \ref{time_comp}. For Selective Search, we use the 'fast-mode' as described in \cite{frcnn}. Our basic HyperNet system takes 1.14 seconds in total, which is 2$\times$ faster than Fast R-CNN. With shared Conv features, the speed up version only takes 20 ms to generate proposals. The total time is 200 ms, which is on par with Faster R-CNN (5 fps) \cite{fasterrcnn}.

\begin{table}[h]\small
\begin{center}
\begin{tabular}{l|cccc}
Method&Conv(shared)&Proposal&Detection&Total\\
\hline
Fast R-CNN&140&2260&170&2570\\
\hline
HyperNet&150&810&180&1140\\
HyperNet-SP&150&20&30&\textbf{200}\\

\end{tabular}
\end{center}
\caption{Timing (ms) on an Nvidia TitanX GPU, except Selective Search proposal is evaluated in a single CPU.}
\label{time_comp}
\vskip -0.2 in
\end{table}

\section{Conclusion}

We have presented HyperNet, a fully trainable deep architecture for joint region proposal generation and object detection. HyperNet provides an efficient combination framework for deep but semantic, intermediate but complementary, and shallow but high-resolution CNN features. A highlight of the proposed architecture is its ability to produce small number of object proposals while guaranteeing high recalls. Both the basic HyperNet and its speed up version achieve state-of-the-art object detection accuracy on standard benchmarks.

\textbf{Acknowledgement} This work was jointly supported by National Natural Science Foundation of China under Grant No. 61210013, 61327809, 91420302 and 91520201.

{\small
\bibliographystyle{ieee}

\begin{thebibliography}{10}\itemsep=-1pt

\bibitem{objectness}
B.~Alexe, T.~Deselaers, and V.~Ferrari.
\newblock What is an object?
\newblock In {\em CVPR}, 2010.

\bibitem{mcg}
P.~Arbelaez, J.~Pont-Tuset, J.~Barron, F.~Marques, and J.~Malik.
\newblock Multiscale combinatorial grouping.
\newblock In {\em CVPR}, 2014.

\bibitem{dpm3}
H.~Azizpour and I.~Laptev.
\newblock Object detection using strongly-supervised deformable part models.
\newblock In {\em ECCV}, 2012.

\bibitem{mincut}
J.~Carreira and C.~Sminchisescu.
\newblock Constrained parametric min-cuts for automatic object segmentation.
\newblock In {\em CVPR}, 2010.

\bibitem{bing}
M.-M. Cheng, Z.~Zhang, W.-Y. Lin, and P.~Torr.
\newblock Bing: Binarized normed gradients for objectness estimation at 300fps.
\newblock In {\em CVPR}, 2014.

\bibitem{dalal2005histograms}
N.~Dalal and B.~Triggs.
\newblock Histograms of oriented gradients for human detection.
\newblock In {\em CVPR}, 2005.

\bibitem{dpm2}
P.~Doll{\'a}r, R.~Appel, S.~Belongie, and P.~Perona.
\newblock Fast feature pyramids for object detection.
\newblock {\em PAMI}, 36(8):1532--1545, 2014.

\bibitem{multibox}
D.~Erhan, C.~Szegedy, A.~Toshev, and D.~Anguelov.
\newblock Scalable object detection using deep neural networks.
\newblock In {\em CVPR}, 2014.

\bibitem{pascalvoc}
M.~Everingham, S.~M.~A. Eslami, L.~Van~Gool, C.~K.~I. Williams, J.~Winn, and
  A.~Zisserman.
\newblock The pascal visual object classes challenge: A retrospective.
\newblock {\em IJCV}, pages 98--136, 2015.

\bibitem{dpm}
P.~F. Felzenszwalb, R.~B. Girshick, D.~McAllester, and D.~Ramanan.
\newblock Object detection with discriminatively trained part-based models.
\newblock {\em PAMI}, 32(9):1627--1645, 2010.

\bibitem{deepboxes}
A.~Ghodrati, M.~Pedersoli, T.~Tuytelaars, A.~Diba, and L.~V. Gool.
\newblock Deepproposal: Hunting objects by cascading deep convolutional layers.
\newblock In {\em ICCV}, 2015.

\bibitem{multiregion}
S.~Gidaris and N.~Komodakis.
\newblock Object detection via a multi-region \& semantic segmentation-aware
  cnn model.
\newblock In {\em ICCV}, 2015.

\bibitem{frcnn}
R.~Girshick.
\newblock Fast r-cnn.
\newblock In {\em ICCV}, 2015.

\bibitem{girshick2014rich}
R.~Girshick, J.~Donahue, T.~Darrell, and J.~Malik.
\newblock Rich feature hierarchies for accurate object detection and semantic
  segmentation.
\newblock In {\em CVPR}, 2014.

\bibitem{hypercol}
B.~Hariharan, P.~Arbel{\'a}ez, R.~Girshick, and J.~Malik.
\newblock Hypercolumns for object segmentation and fine-grained localization.
\newblock In {\em CVPR}, 2015.

\bibitem{sppnet}
K.~He, X.~Zhang, S.~Ren, and J.~Sun.
\newblock Spatial pyramid pooling in deep convolutional networks for visual
  recognition.
\newblock In {\em ECCV}, 2014.

\bibitem{whatobjectness}
J.~Hosang, R.~Benenson, P.~Doll{\'a}r, and B.~Schiele.
\newblock What makes for effective detection proposals?
\newblock {\em PAMI}, 2015.

\bibitem{tracking}
Y.~Hua, K.~Alahari, and C.~Schmid.
\newblock Online object tracking with proposal selection.
\newblock In {\em ICCV}, 2015.

\bibitem{saliency}
Y.~Jia and M.~Han.
\newblock Category-independent object-level saliency detection.
\newblock In {\em ICCV}, 2013.

\bibitem{caffe}
Y.~Jia, E.~Shelhamer, J.~Donahue, S.~Karayev, J.~Long, R.~B. Girshick,
  S.~Guadarrama, and T.~Darrell.
\newblock Caffe: Convolutional architecture for fast feature embedding.
\newblock In {\em ACM Multimedia}, 2014.

\bibitem{alexnet}
A.~Krizhevsky, I.~Sutskever, and G.~E. Hinton.
\newblock Imagenet classification with deep convolutional neural networks.
\newblock In {\em NIPS}, 2012.

\bibitem{deepbox}
W.~Kuo, B.~Hariharan, and J.~Malik.
\newblock Deepbox: Learning objectness with convolutional networks.
\newblock In {\em ICCV}, 2015.

\bibitem{midcut}
T.~Lee, S.~Fidler, and S.~Dickinson.
\newblock Learning to combine mid-level cues for object proposal generation.
\newblock In {\em ICCV}, 2015.

\bibitem{fcn}
J.~Long, E.~Shelhamer, and T.~Darrell.
\newblock Fully convolutional networks for semantic segmentation.
\newblock In {\em CVPR}, 2015.

\bibitem{sift}
D.~G. Lowe.
\newblock Distinctive image features from scale-invariant keypoints.
\newblock {\em IJCV}, 60(2):91--110, 2004.

\bibitem{dpm1}
B.~Pepikj, M.~Stark, P.~Gehler, and B.~Schiele.
\newblock Occlusion patterns for object class detection.
\newblock In {\em CVPR}, 2013.

\bibitem{yolo}
J.~Redmon, S.~Divvala, R.~Girshick, and A.~Farhadi.
\newblock You only look once: Unified, real-time object detection.
\newblock {\em arXiv preprint arXiv:1506.02640}, 2015.

\bibitem{fasterrcnn}
S.~Ren, K.~He, R.~Girshick, and J.~Sun.
\newblock Faster r-cnn: Towards real-time object detection with region proposal
  networks.
\newblock In {\em NIPS}, 2015.

\bibitem{noc}
S.~Ren, K.~He, R.~Girshick, X.~Zhang, and J.~Sun.
\newblock Object detection networks on convolutional feature maps.
\newblock {\em arXiv preprint arXiv:1504.06066}, 2015.

\bibitem{imagenet}
O.~Russakovsky, J.~Deng, H.~Su, J.~Krause, S.~Satheesh, S.~Ma, Z.~Huang,
  A.~Karpathy, A.~Khosla, M.~Bernstein, A.~C. Berg, and L.~Fei-Fei.
\newblock {ImageNet Large Scale Visual Recognition Challenge}.
\newblock {\em IJCV}, pages 1--42, 2015.

\bibitem{overfeat}
P.~Sermanet, D.~Eigen, X.~Zhang, M.~Mathieu, R.~Fergus, and Y.~LeCun.
\newblock Overfeat: Integrated recognition, localization and detection using
  convolutional networks.
\newblock In {\em ICLR}, 2014.

\bibitem{simonyan2014very}
K.~Simonyan and A.~Zisserman.
\newblock Very deep convolutional networks for large-scale image recognition.
\newblock In {\em ICLR}, 2015.

\bibitem{van2011segmentation}
K.~E. Van~de Sande, J.~R. Uijlings, T.~Gevers, and A.~W. Smeulders.
\newblock Segmentation as selective search for object recognition.
\newblock In {\em ICCV}, 2011.

\bibitem{hog}
X.~Wang, T.~X. Han, and S.~Yan.
\newblock An hog-lbp human detector with partial occlusion handling.
\newblock In {\em CVPR}, 2009.

\bibitem{bayesian}
Y.~Zhang, K.~Sohn, R.~Villegas, G.~Pan, and H.~Lee.
\newblock Improving object detection with deep convolutional networks via
  bayesian optimization and structured prediction.
\newblock In {\em CVPR}, 2014.

\bibitem{edgeboxes}
C.~L. Zitnick and P.~Doll{\'a}r.
\newblock Edge boxes: Locating object proposals from edges.
\newblock In {\em ECCV}, 2014.

\end{thebibliography}

}

\end{document}